\documentclass[letterpaper, 10 pt, conference]{ieeeconf}  

\usepackage{cite}

\IEEEoverridecommandlockouts                              

\usepackage{graphics} 
\usepackage{epsfig} 
\usepackage{times} 
\usepackage{amsmath} 
\usepackage{amssymb}  
\usepackage{float}
\usepackage[version=4]{mhchem}
\usepackage{xcolor}

\newtheorem{theorem}{Theorem}

\newtheorem{proposition}{Proposition}

\newtheorem{assumption}{Assumption}
\newtheorem{definition}{Definition}

\title{\LARGE \bf
Learning Genetic Circuit Modules with Neural Networks}

\author{Jichi Wang$^{1}$, Eduardo D. Sontag$^{2}$, and Domitilla Del Vecchio$^{3}$
\thanks{*This work is supported by AFOSR MURI Award Number FA9550-22-1-0316}
\thanks{$^{1}$Department of Mechanical Engineering, Massachusetts Institute of Technology, 77 Massachusetts Avenue, Cambridge, MA 02139, USA.
        {Email: \tt\small jichi@mit.edu}}%
\thanks{$^{2}$Department of Electrical and Computer Engineering and Department of Bioengineering, Northeastern University, Boston, Massachusetts 02115, USA.
        {Email: \tt\small sontag@gmail.com}}%
\thanks{$^{3}$Department of Mechanical Engineering and   Biological Engineering, Massachusetts Institute of Technology, 77 Massachusetts Avenue, Cambridge, MA 02139, USA.
        {Email: \tt\small ddv@mit.edu}}%
}

\begin{document}

\maketitle
\thispagestyle{empty}
\pagestyle{empty}

\begin{abstract}
 In several applications, including in synthetic biology, one often has input/output data on a system   composed of many modules, and although the modules' input/output functions and signals may be unknown, knowledge of the composition architecture can allow to significantly reduce the amount of training data required to learn the system's input/output mapping. Learning the modules' input/output functions is also necessary for designing new systems from different composition architectures. 
 %
 Here, we propose a modular learning framework, which incorporates prior knowledge of the system's compositional structure to (a) identify the composing modules' input/output functions from the system's input/output data and (b) achieve this by using a reduced amount of   data compared to what would be required without knowledge of the compositional structure. To achieve this, we introduce the notion of \textit{modular identifiability}, which allows to recover the modules' input/output functions from a subset of the system's input/output data, and provide theoretical guarantees on a class of systems motivated by  genetic circuits. We illustrate the theory through computational studies, showing that a neural network (NNET) that accounts for the compositional structure is able to learn the composing modules' input/output functions and to predict the system's output on inputs which lie outside of the training set. 
 By reducing the need for experimental data, and allowing modules' identification, this framework offers the potential to ease the design of synthetic biological circuits and of multi-module systems more generally.
\end{abstract}

\section{Introduction} 

In synthetic biology, genetic circuits are commonly designed in a modular fashion, with each genetic module performing specific functions in isolation. However, when these modules are composed together in the cell, their performance can be significantly impacted by interactions with other modules, due to loading effects and resource competition between the modules \cite{jayanthi2013retroactivity, grunberg2020modular, mishra2014loaddriver, delvecchio2008modular, qian2017resource, diblasi2024resourcecomp, diblasi2023resourceaware}. These inter-dependencies complicate the characterization and prediction of a system's behavior. Researchers have been developing physics-based models to help design larger systems while accounting for or mitigating these context effects \cite{delvecchio2015modularity, mishra2014loaddriver, qian2017resource, gyorgy2015isocost, huang2018quasiintegral, jones2020feedforward, frei2020mitigation}, as well as software tools \cite{pandey2023bioscrape, poole2022biocrnpyler} to model genetic circuits with varying complexity of context descriptions. Yet, when composing genetic modules within the cellular host, there remain interactions that are difficult to model through first principles. Therefore, machine learning (ML) models, such as (recurrent) neural networks (NNETs) have been proposed to reduce the uncertainty of physics-based models of genetic circuits \cite{darabi2025combining, palacios2025machine}, in-line with Physics-Informed Neural Networks (PINNs) approaches pioneered in \cite{karniadakis2021physicsinformed}.

These ML approaches applied to genetic circuit modeling, however, require large amounts of data. Also, they focus on identifying a mapping between the inputs and the outputs of a system composed of many genetic modules. Hence, they provide little information about the input/output mapping of the individual genetic modules. This information, in turn, would be useful for composing the modules in new arrangements to achieve new designs. Additionally, accounting for the composition architecture in an ML model may allow the use of substantially less training data to achieve the same predictive ability. Specifically, in the case of a system with multiple inputs, one may be able to train the ML model by activating one input at a time as opposed to requiring data generated from all combinations of inputs. If possible, this would simplify experiments and make data generation faster and cheaper.

In this paper, using information on the compositional structure, we propose a modular learning framework to identify the modules' input/output functions from training data, under certain assumptions on the model architecture. Specifically, we assume that the number of modules is known a priori, and that each module has an unknown input–output function. The outputs of these modules are composed through a partially known map, referred to as the composition map, whose structure is known but whose parameters are unknown. We investigate conditions under which one can learn the modules' input/output functions from input/output data of the system, when the modules' inputs are only activated one at a time. 

We demonstrate that, under \textit{modular identifiability} conditions, this is possible. We thus show that using a NNET architecture that preserves the structure of the composition map, we can learn the modules' input/output functions and predict the global output for arbitrary combinations of inputs when the training set only considers one input being activated at a time. When using the same training data, a ``monolithic'' NNET that does not leverage the structure of the composition is unable to generalize the output prediction on arbitrary input combinations.

\textbf{Related Work.} The idea of training ML models on input/output data where inputs are turned on one at a time to then predict the output on a combination of inputs has been applied before in different synthetic biology settings, with promising results \cite{eslami2022prediction, alcantar2024highthroughput}. However, these works did not provide conditions under which this is possible, which is one of the contributions of this paper.

In the context of networked dynamical systems, a related problem is to estimate local subsystem dynamics when the network topology is known. Under the assumptions that all the subsystems' output signals are linearly summed and some internal signals are directly measurable, the identifiability of the subsystems' LTI transfer functions has been addressed in \cite{vandenhof2013consistentmodule, dankers2016predictorinput}. Related ideas appear in realization theory, where rational linear input/output behaviors can be realized as finite interconnections of components (e.g., integrators, delays, or copies of a fixed subsystem), and an algorithm is provided to determine whether a desired behavior can be implemented using such components~\cite{sontag1979finitary}. In the static case, closer to our problem formulation, identifiability conditions have been provided for subsystems' input/output nonlinear functions under the assumption that these are linearly composed \cite{vizuete2023nonlinear, vizuete2024nonlinear}. In our paper, instead, motivated by the composition architecture found in genetic circuits, we consider nonlinear composition of static nonlinear functions. 

\section{Problem Formulation}
\subsection{General System Formulation}
\begin{figure}[H]
    \centering
    \includegraphics[width=1\linewidth]{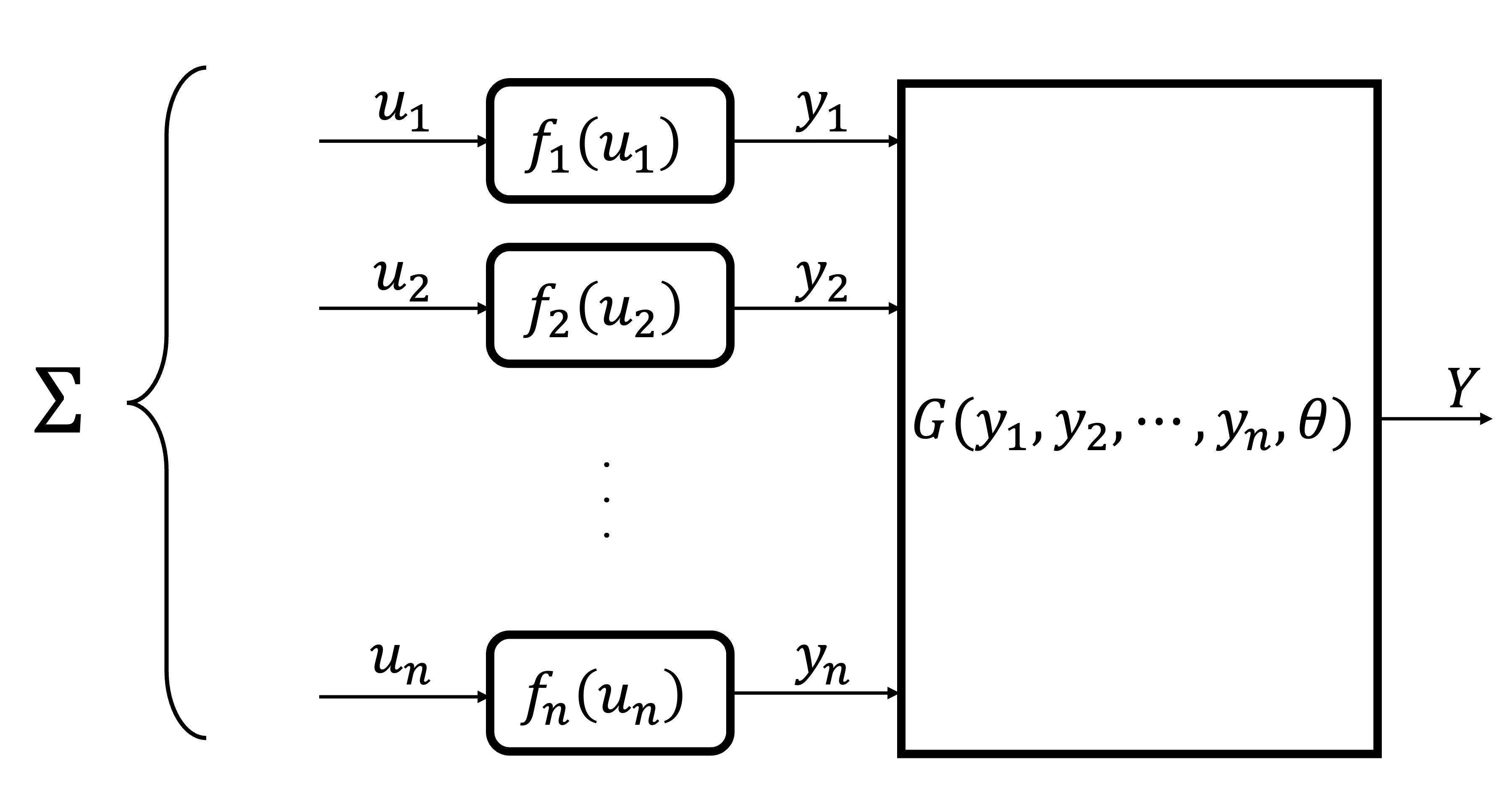}
    \caption{Illustration of the system \(\Sigma\). The system consists of \(n\) modules, each defined by \(y_i = f_i(u_i)\) for \(i \in \{1, 2\}\), whose outputs are fed into a composition map \(G\).}
    \label{fig:system-intro}
\end{figure}

Consider a system \(\Sigma\) composed of \(n\) subsystems or modules. Each module is described by an unknown scalar input/output function \(y_i = f_i(u_i)\), where \(u_i \in \mathbb{R}\) denotes the input to the \(i\)\textsuperscript{th} module. The global measured output \(Y \in \mathbb{R}^n\) is given by a function \(G\colon \mathbb{R}^n \to \mathbb{R}^m\), according to \(Y=G(y_1, y_2, \dots, y_n, \theta)\), in which \(\theta\in \mathbb{R}^p\) is a parameter vector. We call this function the ``composition map''. 
The system structure is illustrated in Fig.~\ref{fig:system-intro}.
In this paper, we will consider the problem of identifying the functions \(f_i\), \(i\in \{1, \dots,n\}\), and parameter \(\theta\) from measurements of \(u = (u_1, \dots, u_n) \in \mathbb{R}^n\) and \(Y\). Specifically, we are interested in solving this problem when the inputs \(u_i\) are varied one at a time. To this end, let $a_i<b_i$ be real numbers, $U_i=[a_i, b_i]\subset \mathbb R$, and $u_i^*\in U_i$. 
Define the uni-modular input set \(\mathcal U\subset \mathbb{R}^n\) as
\begin{equation}
    \mathcal{U} \;:=\; \bigcup_{i=1}^{n} \mathcal{U}_i,
    \label{eqn:U}
\end{equation}
where
\begin{equation}
    \mathcal{U}_i \;:=\;
    \left\{u \in \mathbb{R}^n | \,u_i \in U_i, u_j = u^{*}_j,\, \forall\, j \neq i \right\}. 
    \label{eqn:U-i}
\end{equation}
Then, we ask when it is possible to identify the functions $f_i$ for all $i\in \{1, \dots, n\}$ from measurements of the output \(Y\) of system $\Sigma$ for  $u \in \mathcal{U}$. 

\begin{definition}
The system \(\Sigma\) is said to be \textit{modularly identifiable} on the uni-modular input set \(\mathcal{U}\) if all functions \(f_i(u_i)\) and parameters \(\theta\) can be uniquely recovered from the measured output \(Y\), that is: 
\begin{equation}
    \begin{gathered}
        G(\hat{f}_1(u_1), \dots , \hat{f}_n(u_n), \hat{\theta}) = G(f_1(u_1), \dots, f_n(u_n), \theta), \\ 
        \forall \; u \in \mathcal{U}
        \implies \\
        \hat{f}_i(u_i) = f_i(u_i), \forall \, u_i \in U_i, i \in \{1, 2, \dots, n\}  \text{, and }\hat{\theta} = \theta.
    \end{gathered}
\end{equation}
\end{definition}
In this work, we assume that \(n=m\), $\theta = (\theta_1,\dots,\theta_n) \in \mathbb{R}^n$ and that the \(i\)\textsuperscript{th} entry of $G(\cdot)$ has the following rational form:
\begin{equation}
    G_i\big(f_1(u_1),\dots,f_n(u_n),\theta\big)
    =
    \frac{\theta_i \cdot f_i(u_i)}
         {1 + \displaystyle\sum_{j=1}^n f_j(u_j)}. 
    \label{eqn:n-module-formula--1}
\end{equation}
The form of this map derives from previously developed and experimentally validated models of gene expression when multiple synthetic genetic modules are concurrently operating in the cell \cite{gyorgy2015isocost, qian2017resource}. In the next section, we demonstrate how this form arises.

\section{Model Structure Derivation}
\label{sec:derivation}
\begin{figure}
    \centering
    \includegraphics[width=1\linewidth]{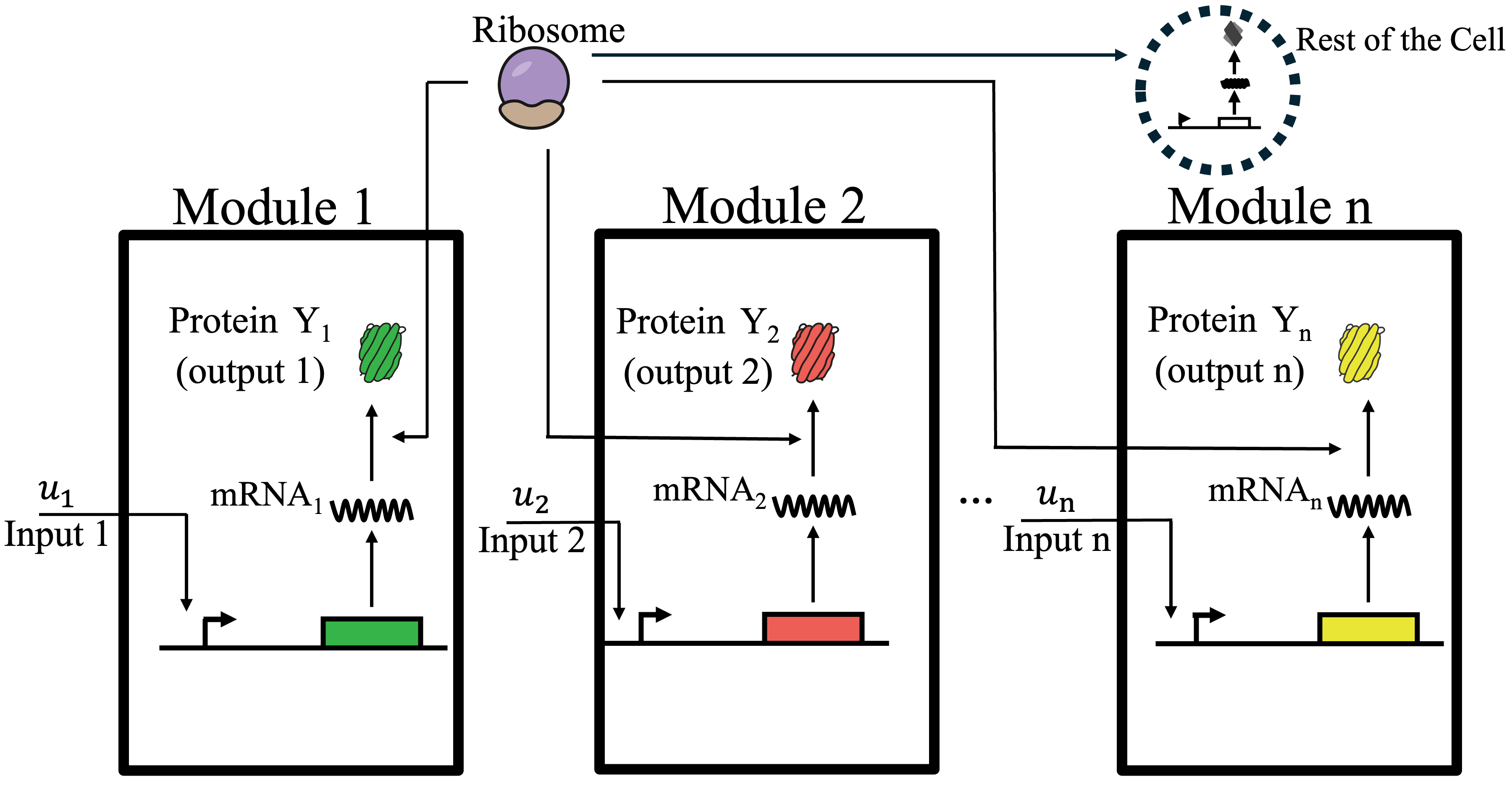}
    \caption{Illustration showing \(n\) transcriptional regulation modules that share ribosomes. }
    \label{fig:n-module-plasmid}
\end{figure}

In this section, we illustrate how the form \eqref{eqn:n-module-formula--1} emerges from the regulation of gene expression. The process of gene expression produces protein from DNA through two steps. In a first step, transcription transforms a DNA sequence into a messenger RNA (mRNA) sequence, and in a second step, translation transforms this mRNA sequence into a sequence of amino acids that then folds into a protein \cite{delvecchio2014biomolecular}. A protein, in turn, can be a transcriptional regulator, activating or repressing the transcription of other genes. These regulators, called transcription factors, can be regarded as the inputs to gene expression modules (Fig.~\ref{fig:n-module-plasmid}). Each such module, can be described by a set of chemical reactions as follows (Chapter 2 of \cite{delvecchio2014biomolecular}). For the \(i\)\textsuperscript{th} module, its process of transcription can be described by 
\begin{equation*}
    \begin{gathered}
        \text{Transcription:} \quad  \ce{\text{DNA}_i ->[\scriptstyle \bar{f}_i(u_i)] \text{mRNA}_i + \text{DNA}_i}, 
    \end{gathered}
\end{equation*}
in which \(\bar{f}_i(u_i)\) is a regulatory function called a Hill function \cite{delvecchio2014biomolecular}, which is increasing with \(u_i\) for an activator or decreasing with \(u_i\) for a repressor. Translation requires a cellular resource called the ribosome (Ribo), which is shared among multiple cellular translation processes, and can be written as: 
\begin{equation*}
    \begin{gathered}
        \text{Translation:} \quad
        \ce{\text{mRNA}_k + \text{Ribo} <=>[a_i][d_i] \text{Ribo:mRNA}_i} \\
        \ce{\text{Ribo:mRNA}_i ->[\scriptstyle k_i] \text{Y}_i + \text{Ribo} + \text{mRNA}_i}, 
    \end{gathered}
\end{equation*}
in which \(\text{Y}_i\) is the protein output of the \(i\)\textsuperscript{th} gene expression module. Protein and mRNA then decay:
\begin{equation*}
    \begin{gathered}
        \ce{\text{Y}_i ->[\scriptstyle \gamma_i] \text{\ensuremath{\varnothing}}}, \quad \ce{\text{mRNA}_i ->[\scriptstyle \delta_i] \text{\ensuremath{\varnothing}}}. 
    \end{gathered}
\end{equation*}
We lump the rest of the cell mRNA into a single species \(\text{mRNA}_{\text{cell}}\) which is produced from \(\text{DNA}_{\text{cell}}\) at a constant rate \(A'\), and free ribosomes reversibly bind with \(\text{mRNA}_{\text{cell}}\) to produce protein \(\text{Y}_{\text{cell}}\), as done elsewhere \cite{delvecchio2014biomolecular, gyorgy2015isocost}:
\begin{equation*}
    \begin{gathered}
        \ce{\text{DNA}_{\text{cell}} ->[\scriptstyle A^{'}] \text{mRNA}_{\text{cell}} + \text{DNA}_{\text{cell}}} \\
        \ce{\text{mRNA}_{\text{cell}} + \text{Ribo} <=>[{a'}][{d'}] \text{Ribo:\(\text{mRNA}_{\text{cell}}\)}} \\
        \ce{\text{Ribo:\(\text{mRNA}_{\text{cell}}\)} ->[\scriptstyle {k'}] \text{Y}_{\text{cell}} + \text{Ribo} + \text{\(\text{mRNA}_{\text{cell}}\)}} \\
        \ce{\text{Y}_{\text{cell}} ->[\scriptstyle {\gamma'}] \text{\ensuremath{\varnothing}}}, \quad
        \ce{\text{mRNA}_{\text{cell}} ->[\scriptstyle {\delta'}] \text{\ensuremath{\varnothing}}}. 
    \end{gathered}
\end{equation*}

The Reaction Rate Equations (RREs) corresponding to the modules' chemical reactions can be written using the law of mass action \cite{delvecchio2014biomolecular} as follows:
\begin{align*}
    &\frac{dY_i}{dt} = k_i \cdot [\text{Ribo:mRNA}_i] - \gamma_i \cdot Y_i, \\[8pt]
    &\frac{d[\text{mRNA}_i]}{dt} =
    \bar{f}_i(u_i) \cdot [\text{DNA}_i]
    - a_i \cdot [\text{mRNA}_i] \cdot [\text{Ribo}] \\[3pt]
    & \hspace{3em}
    + d_i \cdot [\text{Ribo:mRNA}_i]  + k_i \cdot [\text{Ribo:mRNA}]
    \\[3pt]
    & \hspace{3em}
    - \delta_i \cdot [\text{mRNA}_i], \\[8pt]
    &\frac{d[\text{Ribo:mRNA}_i]}{dt} =
    a_i \cdot [\text{mRNA}_i] \cdot [\text{Ribo}] \\[3pt]
    & \hspace{3em}
    - d_i \cdot [\text{Ribo:mRNA}_i]
    - k_i \cdot [\text{Ribo:mRNA}_i].
\end{align*}
Similarly, in the rest of the cell, we have the RREs as follows:
\begin{align*}
&\frac{dY_{\text{cell}}}{dt}
= k' \cdot [\text{Ribo:\(\text{mRNA}_{\text{cell}}\)}]
- \gamma' \cdot Y_{\text{cell}}, \\[8pt]
&\frac{d[\text{mRNA}_{\text{cell}}]}{dt}
= A' \cdot [\text{DNA}_{\text{cell}}]
- a' \cdot [\text{mRNA}_{\text{cell}}] \cdot [\text{Ribo}]
\\[3pt]
&\hspace{3em}
+ d' \cdot [\text{Ribo:\(\text{mRNA}_{\text{cell}}\)}] 
+ k' \cdot [\text{Ribo:\(\text{mRNA}_{\text{cell}}\)}]
\\[3pt]
&\hspace{3em}
- \delta' \cdot [\text{mRNA}_{\text{cell}}], \\[8pt]
&\frac{d[\text{Ribo:\(\text{mRNA}_{\text{cell}}\)}]}{dt}
= a' \cdot [\text{mRNA}_{\text{cell}}] \cdot [\text{Ribo}] \\[3pt]
&\hspace{3em}
- d' \cdot [\text{Ribo:\(\text{mRNA}_{\text{cell}}\)}]
- k' \cdot [\text{Ribo:\(\text{mRNA}_{\text{cell}}\)}].
\end{align*}
Let \(R_T\) be the total concentration of ribosomes, then we have the conservation law:
\[
R_T = [\text{Ribo}] + \sum_{j=1}^n[\text{Ribo:mRNA}_j] + [\text{Ribo:\(\text{mRNA}_{\text{cell}}\)}]. 
\]
The binding and unbinding reactions are much faster than the catalytic reactions and gene expression, i.e., for each \(i\), \(R_T \cdot a_i, d_i \gg k_i, \gamma_i, \delta_i, \bar{f_i}(u_i)\) and \(R_T \cdot a', d' \gg k', \gamma', \delta', A'\) \cite{delvecchio2014biomolecular}. Therefore, we can use the Quasi-Steady-State Approximation (QSSA) to set
\begin{align*}
    \frac{d[\text{Ribo:mRNA}_i]}{dt} = 0,
    \quad
    \frac{d[\text{Ribo:\(\text{mRNA}_{\text{cell}}\)}]}{dt} = 0. 
\end{align*}
Therefore, we have
\[
[\text{Ribo:mRNA}_i] = \frac{[\text{mRNA}_i]}{K_i} \cdot [\text{Ribo}], \quad K_i = \frac{d_i + k_i}{a_i},
\]
\[
[\text{Ribo:\(\text{mRNA}_{\text{cell}}\)}] = \frac{[\text{mRNA}_{\text{cell}}]}{{K_{\text{cell}}}} \cdot [\text{Ribo}], \, {K_{\text{cell}}} = \frac{{d'} + {k'}}{{a'}}. 
\]
Solving for the free ribosome using the conservation law for ribosome, we have that
\[
[\text{Ribo}] = \frac{R_T}{1 + \frac{[\text{mRNA}_{\text{cell}}]}{K_{\text{cell}}} + \sum_{j = 1}^n\frac{[\text{mRNA}_j]}{K_j}}. 
\]
Then, for each \(i\), we arrive at simplified differential equations for \(\frac{dY_i}{dt}\), \(\frac{d[\text{mRNA}_i]}{dt}\), and \(\frac{d[\text{mRNA}_{\text{cell}}]}{dt}\):
\begin{align*}
    &\frac{dY_i}{dt} = \frac{k_i [\text{mRNA}_i]}{K_i} \cdot \frac{R_T}{1 + \frac{[\text{mRNA}_{\text{cell}}]}{K_{\text{cell}}} + \sum_{j = 1}^n \frac{[\text{mRNA}_j]}{K_j}} - \gamma_i Y_i, \\
    &\frac{d[\text{mRNA}_i]}{dt} = \bar{f}_i(u_i) \cdot [\text{DNA}_i] - \delta_i \cdot [\text{mRNA}_i], \\
    &\frac{d[\text{mRNA}_{\text{cell}}]}{dt} = A' \cdot [\text{DNA}_{\text{cell}}] - \delta' \cdot [\text{mRNA}_{\text{cell}}], 
\end{align*}
which is consistent with standard models \cite{qian2017resource}. At the unique equilibrium of the system, the Jacobian matrix is upper triangular with negative diagonal entries. Consequently, all eigenvalues are negative and the equilibrium is locally exponentially stable. Since we are interested in the steady state input/output mapping of the system, we then set the derivatives to zero to obtain the equilibrium and hence the global input/output function:
\[
Y_i = \frac{\theta_i \cdot f_i(u_i)}{1 + \sum_{j=1}^n f_j(u_j)}, \qquad i \in \{1, 2\}, 
\]
in which 
\begin{align*}
    &\theta_i = \frac{R_T  \, k_i}{\gamma_i}, \\
    &f_i(u_i) = \frac{[\text{DNA}_i]}{K_i \cdot \delta_i} \cdot \frac{1}{1 + \frac{A' \cdot [\text{DNA}_{\text{cell}}]}{K_{\text{cell}} \cdot \delta'}} \cdot \bar f_i(u_i). 
\end{align*}
This coincides with \eqref{eqn:n-module-formula--1}. 

\subsection{Modular Identifiability when \(\theta = 1\)}
\label{sec:3b}
Here, we assume \(\theta = 1\). Let \(f, \hat{f}\colon [0, 1] \to \mathbb{R}\), and \(f(u), \hat{f}(u) \geq 0\) for \(u \in [0, 1]\), where $\hat f(u)$ is used to approximate $f(u)$. Let
\[
G(f(u)) = \frac{f(u)}{1 + f(u)}. 
\]
\begin{proposition}
The model is \textit{modularly identifiable}, i.e., for all \(u \in [0, 1]\),
\begin{equation*}
    G(f(u)) = G(\hat{f}(u)) \implies f(u) = \hat{f}(u). 
\end{equation*}
\end{proposition}
\begin{proof}
    Since the function \(G(s) = \frac{s}{1 + s}\) is strictly increasing for \(s \geq 0\), it is injective on its domain \(\mathbb{R}^+\). By assumption, \(f(u), \hat{f}(u)\) are positive for all \(u \in [0, 1]\). Thus, if \(G(f(u)) = G(\hat{f}(u))\) for all \(u \in [0, 1]\), then \(f(u) = \hat{f}(u)\) for all \(u \in [0, 1]\). 
\end{proof}

\vspace{\baselineskip}

\begin{proposition}
    For any \(\epsilon > 0\), there exists a \(\delta > 0\), such that
\begin{multline*}
    \left |G(f(u)) - G(\hat{f}(u))\right | < \delta \\
    \implies \left |f(u) - \hat{f}(u)\right | < \epsilon \text{, for } u \in [0, 1]. 
\end{multline*}
\end{proposition}
\begin{proof}
    Since \(G(s)\), \(f(u)\), and \(\hat{f}(u)\) are continuous, the sets \(G(f([0,1]))\) and \(G(\hat{f}([0,1]))\) are compact. Hence, their union \(K := G(f([0,1])) \cup G(\hat{f}([0,1]))\) is a compact subset of \([0,1)\). Additionally, as \(G(s)\) is injective, it has an inverse, given by 
\[
G^{-1}(y) = \frac{y}{1 - y},\; y \in K. 
\]
On the compact set \(K\), the derivative 
\[
(G^{-1})'(y) = \frac{1}{(1 -y)^2}
\]
is continuous and thus bounded. This implies that \(G^{-1}\) is Lipschitz on \(K\) with Lipschitz constant \(L > 0\). We thus have that
\begin{multline*}
    \left |f(u) - \hat{f}(u)\right | = \left | G^{-1}(G(f(u))) - G^{-1}(G(\hat{f}(u)))\right | \\
    \leq L \left |G(f(u)) - G(\hat{f}(u))\right |. 
\end{multline*}
Thus, given any \(\epsilon > 0\), by picking \(\delta = \frac{\epsilon}{L}\), we prove the statement. 
\end{proof}

\vspace{\baselineskip}

Proposition 2 implies that if \(G(f(u))\) can be estimated with sufficiently high accuracy, then \(f(u)\) will likewise be estimated with comparable accuracy. We will illustrate this point in the next section with an example.


\subsection{Example}
\begin{figure}
    \centering
    \includegraphics[width=1\linewidth]{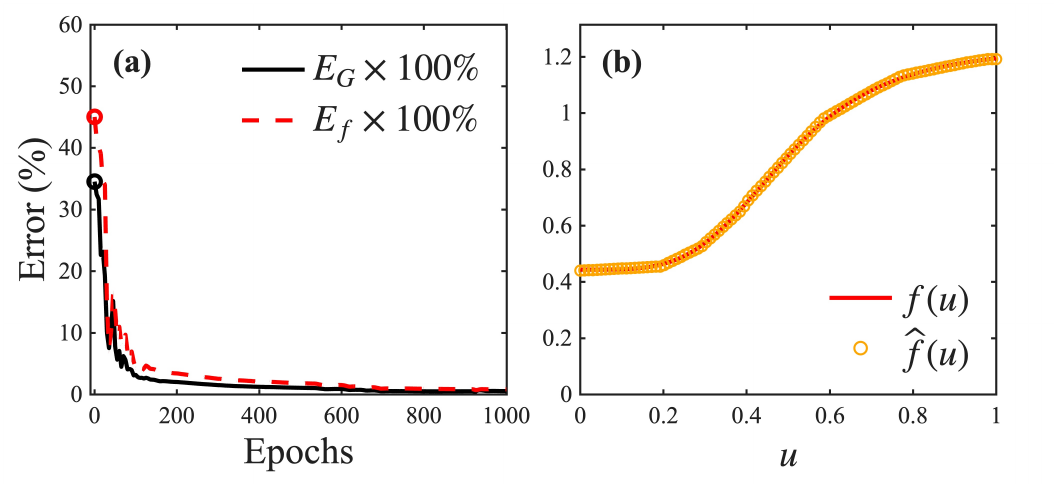}
    \caption{ Define \(E_G := \max_k |G(\hat f(u_k)) - G(f (u_k))|/ \max_k G(f(u_k))\) and \(E_f := \max_k |\hat f(u_k) - f(u_k)| / \max_k f(u_k)\), in which \(u_k\) are the elements of the training set. (a) The plot shows the error convergence over 1000 epochs. The training dataset comprises 100 input/output pairs \((u_k, Y_k)\), where inputs \(u_k\) is uniformly sampled from the interval \([0, 1]\). We choose \(\hat{f}(u)\) as a fully connected feedforward NNET with four hidden layers, each containing 20 ReLU-activated neurons, and initialize its weights using Kaiming initialization \cite{he2015delving}. The NNET \( \hat{f}(u) \) is trained using the Adam optimizer with a learning rate of 0.1 and full-batch gradient descent, with the cost function given by the mean square error \(\frac{1}{100}\sum_{k = 1}^{100} |G(\hat f(u_k)) - G(f(u_k))|^2\). (b) The plot shows the comparison between the trained \( \hat{f}(u) \) and true \( f(u) \) at the final (1000\textsuperscript{th}) epoch.}
    \label{fig:single_module_error_convergence_combined}
\end{figure}

Here, we assume to have input/output data $(u,Y)$ for \(Y=G(f(u))\) and $u\in [0,1]$ as training set, and use a ML model given by:
\[
G(\hat{f}(u)) = \frac{\hat{f}(u)}{1 + \hat{f}(u)}, 
\]
in which \(\hat{f}(u)\) is a NNET. According to Proposition 2, if we make the error \(|G(\hat{f}(u)) - G(f(u))|\) sufficiently small, we should also be making the error \(|\hat{f}(u) - f(u)|\) sufficiently small. As an example, we let
\[
    f(u) = \frac{0.797 \cdot \left(\frac{u}{0.494}\right)^4}{1 + \left(\frac{u}{0.494}\right)^4} + 0.443.
\]
As shown in Fig.~\ref{fig:single_module_error_convergence_combined}(a), \(|\hat{f}(u) - f(u) |\) converges to 0, as \(|G(\hat{f}(u)) - G(f(u)) |\) converges to 0, which verifies Proposition 2. Fig.~\ref{fig:single_module_error_convergence_combined}(b) shows the trained \(\hat{f}(u)\) against the true \(f(u)\) at the end epoch. 

\subsection{Modular Identifiability for Unknown \(\theta\)}
If $\theta$ is unknown and $f(u), \hat f(u)$ are generic real-valued functions, then $\theta$ and $f(u)$ cannot be identified from a single activation of the module alone. As an example, suppose that \(\theta \in \mathbb{R}\) is unknown 
and \(f(u)\) is a Hill function:
\[
f(u) = \frac{u}{1 + u}, 
\]
then any \(\hat{\theta} \neq \theta\) and \(\hat{f}(u) = \frac{\frac{\theta}{\hat{\theta}} \cdot f(u)}{1 + \frac{\hat{\theta}- \theta}{\hat \theta} \cdot f(u)}\) will give \(G(\hat{f}(u), \hat{\theta}) = G(f(u), \theta)\) but \(\hat{f}(u) \neq f(u)\). 


\section{Case of Multiple Modules}
Without loss of generality, we consider two modules to simplify notation, that is, \(n=2\). Modular identifiability for the case of \(n>2\) modules can be proved in a similar way. Let
\begin{gather*}
\mathcal{U}_1=\{u\in \mathbb R^2 \;|\; u_1\in [0,1], u_2=1\}, \\
\mathcal{U}_2=\{u\in \mathbb R^2 \;|\; u_1 = 1, u_2 \in [0, 1]\}, \\
\mathcal{U} = \mathcal{U}_1 \cup \mathcal{U}_2. 
\end{gather*}
From \eqref{eqn:n-module-formula--1}, the measured output \(Y\) can be expressed as:
\begin{equation}
    Y = G(f_1(u_1), f_2(u_2), \theta) = 
    \begin{bmatrix}
        G_1(f_1(u_1), f_2(u_2), \theta)\\
    G_2(f_1(u_1), f_2(u_2), \theta) \label{eqn:2-module-formula--1}
    \end{bmatrix},
\end{equation}
with \(\theta = (\theta_1, \theta_2) \in \mathbb{R}^2\) and 
\begin{equation}
    G_i(f_1(u_1), f_2(u_2), \theta)= \frac{\theta_i \cdot f_i(u_i)}{1 + f_1(u_1) + f_2(u_2)}, \;i\in\{1,2\}.
    \label{eqn:2-module-formula--2}
\end{equation}
Here, we make the following assumption. 
\begin{assumption}
Let $f_1, f_2 : [0,1] \to \mathbb{R}$ be non-constant continuous functions satisfying 
$f_1(1) \neq 0$ and $f_2(1) \neq 0$. 
This condition is consistent with typical biological settings, where basal expression leads to nonzero output at any input level. 
\end{assumption}

\subsection{Motivation: Regulation of Two Gene Expression Modules}
In this case, we consider two transcriptional modules that are coupled due to competition for a shared cellular resource, the ribosome.

\subsection{Modular Identifiability}
In this section, we give the main result of the paper, conditions that ensure we can estimate the functions \(f_i(u_i)\), \(i\in \{1,2\}\), from measurements of \(Y\) on the input set  \(\mathcal{U}\) which varies the input for one module at a time. 

\begin{theorem}
    The model in \eqref{eqn:2-module-formula--1}-\eqref{eqn:2-module-formula--2} is \textit{modularly identifiable}, that is:
\begin{gather*}
    G_i(\hat{f}_1(u_1), \hat{f}_2(u_2), \hat{\theta}) = G_i(f_1(u_1), f_2(u_2), \theta),\\
    \forall \, u \in \mathcal{U}, \forall \, i \in \{1, 2\}
    \\ \implies
    \hat{f}_i(u_i) = f_i(u_i) \text{ and } \hat{\theta}_i = \theta_i, \forall \,i \in \{1, 2\}, \forall \, u_i\in [0, 1]. 
\end{gather*}
\end{theorem}
\begin{proof}
Define the following constants:
\[
A_1 = f_1(1), \, \hat{A}_1 = \hat{f}_1(1),
A_2 = f_2(1), \, \hat{A}_2 = \hat{f}_2(1).
\]
Also, define the following functions:

\(G_1(f_1(u_1), f_2(u_2), \theta)\) for $(u_1,u_2)\in \mathcal U_1$ as:
\[
G_{11}(u_1) = \frac{\theta_1 \cdot f_1(u_1)}{1 + f_1(u_1) + A_2}, u_1 \in [0, 1]
\]

\(G_2(f_1(u_1), f_2(u_2), \theta)\) for $(u_1,u_2)\in \mathcal U_1$ as:
\[
G_{21}(u_1) = \frac{\theta_2 \cdot A_2}{1 + f_1(u_1) + A_2}, u_1 \in [0, 1]
\]

\(G_1(f_1(u_1), f_2(u_2), \theta)\) for $(u_1,u_2)\in \mathcal U_2$ as:
\[
G_{12}(u_2) = \frac{\theta_1 \cdot A_1}{1 + A_1 + f_2(u_2)}, u_2 \in [0, 1]
\]

\(G_2(f_1(u_1), f_2(u_2), \theta)\) for $(u_1,u_2)\in \mathcal U_2$ as:
\[
G_{22}(u_2) = \frac{\theta_2 \cdot f_2(u_2)}{1 + A_1 + f_2(u_2)}, u_2 \in [0, 1].
\]
Let $\hat{G}_{ij}(u_j)$ for $i,j\in \{1,2\}$ be defined as $G_{ij}(u_j)$ by replacing $\theta_i$ with $\hat\theta_i$, $A_i$ with $\hat A_i$, and $f_j(u_j)$ with $\hat f_j(u_j)$. We then have
\begin{multline}
    G_{11}(u_1) - \hat{G}_{11}(u_1) = 0 \implies \\
    \theta_1 \cdot f_1(u_1) \cdot (1 + \hat{A}_2) - \hat{\theta}_1 \cdot \hat{f}_1(u_1) \cdot (1 + A_2) + \\
    (\theta_1 - \hat{\theta}_1) \cdot f_1(u_1) \cdot \hat{f}_1(u_1) = 0, \label{G11-G11hat}
\end{multline}
and 
\begin{multline}
    G_{21}(u_1) - \hat{G}_{21}(u_1) = 0 \implies \\
    \theta_2 \cdot A_2 \cdot (1 + \hat{f}_1(u_1)) - \hat{\theta}_2 \cdot \hat{A}_2 \cdot (1 + f_1(u_1)) + \\
    (\theta_2 - \hat{\theta}_2) \cdot A_2 \cdot \hat{A}_2 = 0 \label{G21-G21hat}. 
\end{multline}
From \eqref{G11-G11hat} and \eqref{G21-G21hat}, we can obtain two expressions for \(\hat{f}(u_1)\). Then, by equating these two expressions and rearranging the terms, we obtain
\begin{equation}
    \alpha \cdot f_1(u_1)^2 + \beta \cdot f_1(u_1) + \gamma = 0, \label{f-poly}
\end{equation}
where 
{\small
\begin{align*}
\alpha &= \tfrac{\hat\theta_2 \hat A_2}{\theta_2 A_2} (\theta_1 - \hat\theta_1), \\[1ex]
\beta  &= (\theta_1 - \hat\theta_1)\tfrac{\hat\theta_2 \hat A_2}{\theta_2 A_2} 
          + \theta_1(1+\hat A_2) \\
       &\quad - \hat\theta_1(1+A_2)\tfrac{\hat\theta_2 \hat A_2}{\theta_2 A_2}
          - (\theta_1 - \hat\theta_1)\Big(\tfrac{\theta_2-\hat\theta_2}{\theta_2}\hat A_2+1\Big), \\[1ex]
\gamma &= -\hat\theta_1(1+A_2)\tfrac{\hat\theta_2 \hat A_2}{\theta_2 A_2} 
          + \hat\theta_1(1+A_2)\Big(\tfrac{\theta_2-\hat\theta_2}{\theta_2}\hat A_2+1\Big).
\end{align*}
}Since \(f_1(u_1)\) is continuous and not a constant, its image \(f_1([0, 1])\), by the Intermediate Value Theorem, is an interval in \(\mathbb{R}\). Therefore, the left-hand side of \eqref{f-poly} is a polynomial in \(f_1(u_1)\) that vanishes on a subset of \(\mathbb{R}\) that has infinitely many points. Then, using the Polynomial Identity Theorem \cite{hungerford1974algebra}, we have that all the coefficients of \(f_1(u_1)^2\), \(f_1(u_1)\), and \(f_1(u_1)^0\) must be 0. Then, setting \(\alpha = 0\) gives \(\hat{\theta}_1 = \theta_1\). Using the same argument for \(G_{12}(u_2)\) and \(G_{22}(u_2)\), we can obtain that \(\hat{\theta}_2 = \theta_2\). Setting \(\beta = 0\) and \(\gamma = 0\) with \(\hat{\theta}_1 = \theta_1\) and \(\hat{\theta}_2 = \theta_2\), we then have 
\[
(1 + \hat{A}_2) - (1 + A_2) \cdot \frac{\hat{A}_2}{A_2} = 0 \implies \hat{A}_2 = A_2. 
\]
Thus, it follows from \eqref{G21-G21hat} that \(\hat{f}_1(u_1) = f_1(u_1)\) for all $u_1\in [0,1]$. Again, applying this argument to  \(G_{12}(u_2)\) and \(G_{22}(u_2)\), we obtain \(\hat{f}_2(u_2) = f_2(u_2)\) for all $u_2\in [0,1]$. 
\end{proof}

\vspace{\baselineskip}

In the next theorem, we show that if the approximating functions \(G_1(\hat{f}_1(u_1), \hat{f}_2(u_2), \hat{\theta})\) and \(G_2(\hat{f}_1(u_1), \hat{f}_2(u_2), \hat{\theta})\) are arbitrarily close to the true functions on the input set \(\mathcal{U}\), then \(\hat{f}_1(u_1), \hat{f}_2(u_2)\) will be close to \(f_1(u_1), f_2(u_2)\). To achieve this, we make the following assumptions. 
\begin{assumption}
    For all \(i \in \{1, 2\}\), let \(f_i, \hat f_i \in \mathcal F_i\), where $\mathcal{F}_i = \{ f \in C([0,1]) : \|f\|_\infty \le M,\; |f(u)-f(v)| \le L|u-v| \text{ for all } u,v \in [0,1],\; |f(1) - f(0)| > \rho, \rho> 0, \; |f(u)| \ge d|u|, \text{ for all } u \in [0,1], d > 0\}$ with some finite \(M, L \in \mathbb{R}\). Furthermore, suppose \(\theta \in \Theta\), where \(\Theta \subset \mathbb{R}^n\) is compact. This automatically implies Assumption~1, since each \(f_i\) is constrained to be strictly positive and constant functions are excluded.
\end{assumption}


\begin{theorem}
\label{theorem:wellposedness-n-modules}
    For any \(\epsilon > 0\), there exists a \(\delta > 0\), such that if 
    \begin{multline*}
            \left|G_i(\hat f_1(u_1), \hat f_2(u_2),\hat\theta)- G_i(f_1(u_1), f_2(u_2),\theta)\right|<\delta, \\
            \forall \;i\in \{1, 2\}, \forall\; (u_1, u_2) \in \mathcal U,
    \end{multline*}
    then
    \[
    \displaystyle \sup_{u_i\in U_i} \left |\hat f_i(u_i) - f_i(u_i)\right| < \epsilon \text{, and } \left|\hat \theta_i - \theta_i \right| < \epsilon, \\ \forall \, i \in \{1, 2\}. 
    \]
\end{theorem}

\begin{proof}
Define the parameter space $\mathcal{X} := \mathcal{F}_1 \times \mathcal{F}_2 \times \Theta$.
For \(x \in \mathcal{X}\), equip \(\mathcal{X}\) with the following norm:
\[
\|x\|_{\mathcal X} = \max \{\|f_1\|_\infty, \|f_2\|_\infty, \|\theta\|_\infty\},
\]
where \(\displaystyle \|f_i\|_{\infty}:= \sup_{u_i\in U_i} |f_i(u_i)|\) for \(i \in \{1, 2\}\), and for two elements $x = (f_1,f_2,\theta)$ and $\hat{x} = (\hat{f}_1, \hat{f}_2,\hat{\theta})$, 
the induced metric is
\[
\|x - \hat{x}\|_{\mathcal X} 
= \max \{\|f_1 - \hat f_1\|_\infty, \|f_2 - \hat f_2\|_\infty, \|\theta - \hat \theta\|_\infty\}.
\]
Let $\bar G: C([0,1])^2 \times \mathbb{R}^2 \to C(\mathcal U,\mathbb{R}^2)$ be the forward map defined by $\bar G(f_1,f_2,\theta)(u_1,u_2):=G(f_1(u_1),f_2(u_2),\theta)$, \((u_1, u_2) \in \mathcal{U}\), and let \(\mathcal{Y} := \bar G(\mathcal{X})\). Then, equip \(\mathcal{Y}\) with
\begin{align*}
\|\bar G(f_1, f_2, &\theta) - \bar G(\hat f_1, \hat f_2, \hat \theta)\|_{\mathcal{Y}} \\
&= \max_{i \in \{1, 2\}} \sup_{(u_1, u_2)\in \mathcal{U}}
\big|\bar G_i(f_1, f_2, \theta)(u_1, u_2) - \\& \bar G_i(\hat f_1, \hat f_2, \hat \theta)(u_1, u_2)\big|.\\
&= \max_{i \in \{1, 2\}} \sup_{(u_1, u_2)\in \mathcal{U}}
\big|G_i(f_1(u_1), f_2(u_2), \theta) - \\ &G_i(\hat f_1(u_1), \hat f_2(u_2), \hat \theta)\big|. 
\end{align*}
We first claim that \(\bar G\) is injective. Suppose
\[
\bar G(f_1,f_2,\theta)=\bar G(\hat f_1,\hat f_2,\hat\theta).
\]
Then, by the definition of \(\bar G\),
\[
G\big(f_1(u_1),f_2(u_2),\theta\big)
=
G\big(\hat f_1(u_1),\hat f_2(u_2),\hat\theta\big), \forall (u_1,u_2)\in\mathcal U.
\]
From Theorem~1, we thus obtain \(f_i(u_i)=\hat f_i(u_i)\) for all \(u_i\in U_i\) and all \(i\), and \(\theta=\hat\theta\). Hence
\[
(f_1,f_2,\theta)=(\hat f_1,\hat f_2,\hat\theta),
\]
so \(\bar G\) is injective.

By the Arzelà--Ascoli Theorem, each $\mathcal{F}_i$ is compact, and thus the product $\mathcal{X}$ is compact. Since $\bar G$ is continuous on \(\mathcal{X}\) and $\mathcal{X}$ is compact, $\mathcal{Y}$ is also compact. From Theorem~1, \(\bar G: \mathcal{X} \to \mathcal{Y}\) is injective, and hence $\bar G$ is a bijection onto its image $\mathcal{Y}$. Thus, by Theorem~26.6 in \cite{munkres2000topology}, a continuous bijection 
$\bar G$ from a compact $\mathcal{X}$ to a Hausdorff space $\mathcal{Y}$ 
is a homeomorphism. As a result, $\bar G^{-1} : \mathcal{Y} \to \mathcal{X}$ is continuous on $\mathcal{Y}$. Furthermore, since $\mathcal{Y}$ is compact, the Heine--Cantor Theorem implies that $\bar G^{-1}$ is uniformly continuous on \(\mathcal{Y}\). 
Therefore, for any $\epsilon > 0$, there exists a $\delta > 0$ such that
\begin{multline*}
   \|\bar G(f_1, f_2, \theta) - \bar G(\hat f_1,\hat f_2,\hat\theta)\|_{\mathcal{Y}} < \delta \\
\implies
\|\bar G^{-1}(\bar G(f_1,f_2,\theta)) - \bar G^{-1}(\bar G(\hat f_1,\hat f_2,\hat\theta))\|_{\mathcal{X}} < \epsilon. 
\end{multline*}
By the definition of the product norm on $\mathcal{X}$, this implies
\[
\displaystyle \sup_{u_i\in U_i}|\hat f_i(u_i) - f_i(u_i)| < \epsilon
\text{, and }
|\theta_i - \hat \theta_i| < \epsilon, \forall\, i\in\{1,2\},
\]
which completes the proof.
\end{proof}

\subsection{Computational Study}
Here, we demonstrate that, by virtue of Theorem 1 and 2, we can learn the subsystems' functions \(f_1, f_2\) and the parameters \(\theta_1, \theta_2\) from measurements of the global output \(Y\) on the input set \(\mathcal{U}\). In particular, we define the following functions and parameters. Let \(f_1(u_1)\) be an activating Hill function and \(f_2(u_2)\) be a repressing Hill function.
\[
f_1(u_1) = \frac{0.326 \cdot \Big (\frac{u_1}{0.952} \Big )^4}{1 + \Big (\frac{u_1}{0.952} \Big )^4} + 0.176, \quad \theta_1 = 0.703,
\]
\[
f_2(u_2) = \frac{0.261}{1 + \Big (\frac{u_2}{0.415} \Big )^2} + 0.192, \quad \theta_2 = 0.204. 
\]
As our ML model, we take \(\hat{f}_1, \hat{f}_2\) as NNETs and \(\hat{\theta}_1, \hat{\theta}_2\) as learnable parameters. Our training dataset consists of 200 pairs \((u_k, Y_k)\) with \(u_k \in \mathcal{U} = \mathcal{U}_1 \cup \mathcal{U}_2\). The first 100 inputs are of the form \((\bar{u}_1, 1)\), where \(\bar{u}_1\) is uniformly sampled from \([0,1]\). Similarly, the second 100 inputs are of the form \((1, \bar{u}_2)\), where \(\bar{u}_2\) is uniformly sampled from \([0,1]\). We optimize the following loss function:
\begin{equation} \label{eqn:squared-error}
    \begin{aligned}
\frac{1}{200}\sum_{k = 1}^{200} &\left|G_1(\hat f(u_{1_k}), \hat f(u_{2_k}), \hat \theta) -  G_1( f(u_{1_k}), f(u_{2_k}), \theta)\right|^2 \\
    + &\left|G_2(\hat f(u_{1_k}), \hat f(u_{2_k}), \hat \theta) -  G_2( f(u_{1_k}), f(u_{2_k}), \theta)\right|^2. 
\end{aligned}
\end{equation}
Here, \( \hat{f}_1(u_1), \hat{f}_2(u_2) \) are chosen as fully connected feedforward NNETs with four hidden layers, each containing 20 ReLU-activated neurons. The weights of \( \hat{f}_1(u_1), \hat{f}_2(u_2) \) are initialized using Kaiming initialization \cite{he2015delving}, and \(\hat{\theta}_1, \hat{\theta}_2\) are initialized as 3. Training is performed using the Adam optimizer with a learning rate of 0.005 and full-batch gradient descent. 

\begin{figure}
    \centering
    \includegraphics[width=1\linewidth]{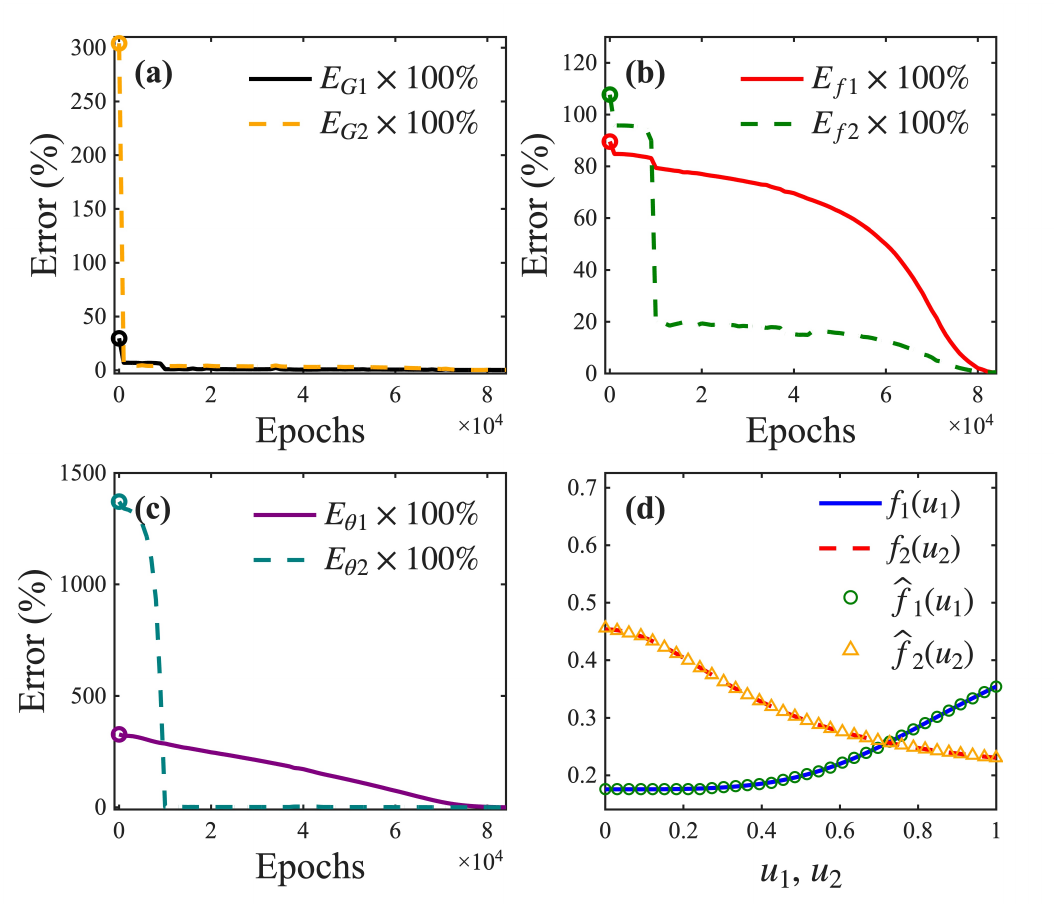}
    \caption{Error convergence over 84{,}000 epochs for the case of two modules. Define the following errors: \(E_{Gi} := \max_{k} |G_i(\hat f_1(u_{1_k}), \hat f_2(u_{2_k}), \hat \theta) - G_i(f_1(u_{1_k}), f_2(u_{2_k}), \theta)| / \max_{k}G_i(f_1(u_{1_k}), f_2(u_{2_k}), \theta)\), \(E_{fi} := \max_{k} |\hat f_i(u_{i_k}) - f_i(u_{i_k})| / \max_{k}f_i(u_{i_k})\), and \(E_{\theta i} := |\hat \theta_i - \theta_i| / \theta_i\), for \(i = \{1, 2\}\). (a) The plot shows the convergence of \(E_{G1}\) and \(E_{G2}\). (b) The plot shows the convergence of \(E_{f1}\) and \(E_{f2}\). (c) The plot shows the convergence of \(E_{\theta 1}\) and \(E_{\theta 2}\). (d) The plot shows the comparison between the true functions \( f_1(u_1) \), \( f_2(u_2) \) and their learned counterparts \( \hat{f}_1(u_1) \), \( \hat{f}_2(u_2) \) at the final (84{,}000\textsuperscript{th}) epoch. Also, at the final epoch, the learned parameters \(\hat{\theta}_1 = 0.70325, \hat{\theta}_2 = 0.20361\) vs. the true parameters \(\theta_1 = 0.70340, \theta_2 = 0.20357\).}
    \label{fig:2_modules_combined_2by2_plot}
\end{figure}

As shown in Fig.~\ref{fig:2_modules_combined_2by2_plot}(a)-(c), for \(i \in \{1, 2\}\), \(|\hat{f}_i(u_i) - f_i(u_i)|\) and \(|\hat \theta_i - \theta_i|\) converge to 0, as \(|G_i(\hat f_1(u_1), \hat f_2(u_2), \hat \theta) - G_i(f_1(u_1), f_2(u_2), \theta)|\), \(i \in \{1, 2\}\), converge to 0, which verifies Theorem 2. The comparison between the learned model \(\hat{f}_1(u_1)\), \(\hat{f}_2(u_2)\) and the true functions \(f_1(u_1)\), \(f_2(u_2)\) at the final epoch is shown in Fig.~\ref{fig:2_modules_combined_2by2_plot}(d).

\begin{figure}
    \centering
    \includegraphics[width=1\linewidth]{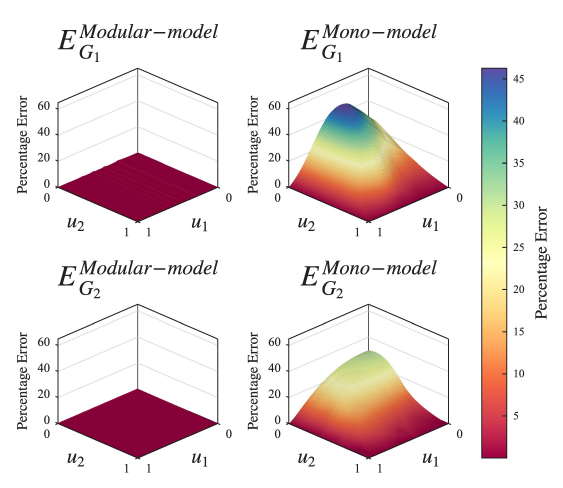}
    \caption{We use the same simulation setup as in Fig~\ref{fig:2_modules_combined_2by2_plot} to compare the generalization ability of our modular learning approach and a monolithic learning approach. The modular learning model is the same as in Fig~\ref{fig:2_modules_combined_2by2_plot}. Let \(\textit{Modular-model}\) denote the trained modular learning model and \(\textit{Mono-model}\) be the trained monolithic learning model. For \(M \in \{\textit{Modular-model}, \textit{Mono-model}\}\) and \(i \in \{1, 2\}\) , the point-wise error is defined as \(E_{G_i}^M(u_1, u_2):= |M(u_1, u_2)_i - G_i(f_1(u_1), f_2(u_2), \theta)| / G_i(f_1(u_1), f_2(u_2), \theta)\). The loss surfaces are generated from 10{,}000 test grid points uniformly sampled over \([0,1]^2\). }
    \label{fig:prediction-3d-loss}
\end{figure}

{\bf Prediction of system's output on out-of-distribution test data.} Because the estimated modules' functions and parameter $\theta$ are very close to the true entities, we expect that our learned model, trained on the input set $\mathcal U$ in which only one input is varied at a time, should predict well the output from arbitrary input combinations. These input combinations form test data that lie outside the distribution of the training data. We therefore evaluate the learned model prediction on this out-of-distribution test data against the predictions of a monolithic NNET trained on the same input set $\mathcal U$. This is shown in Fig~\ref{fig:prediction-3d-loss}. The monolithic ML model is taken to be a fully connected feedforward NNET with two inputs, two outputs, and four hidden layers of 50 ReLU-activated neurons each, with the weights initialized using Kaiming initialization. We used the Adam optimizer with a learning rate of \(0.001\) to minimize the same error in \eqref{eqn:squared-error}, training the model for 8{,}000 epochs until the training error converged.

From these plots, we see that while the modular learning model generalizes well on the entire input set, the monolithic approach does not generalize well for arbitrary combinations of inputs, but only for those input combinations that belong to the distribution of the training set.

\section{Conclusion}
In this paper, we introduced a modular ML framework, which leverages prior knowledge of the composition structure of a system to learn the input/output functions of the composing modules from input/output data of the system. We have also demonstrated how, by learning the modules, the modular ML model can generalize on out-of-distribution data, whereas a monolithic ML model fails to do so. 

As a running example, we considered a system architecture that emerges when composing genetic circuit modules in the cell, which share cellular resources. In this setup, we have considered an example involving several input/output genetic modules operating in the cell and competing for ribosomes. We have demonstrated that it is sufficient to activate one module at a time to enable predictions on arbitrary combinations of modules' inputs. From a practical point of view, this allows to dramatically reduce the data requirement for enabling prediction of the behavior of many modules operating together in the cell while accounting for the effects of context-dependence \cite{jayanthi2013retroactivity, grunberg2020modular, mishra2014loaddriver, delvecchio2008modular, qian2017resource, diblasi2024resourcecomp, diblasi2023resourceaware}.

In our work, since we assume no prior knowledge of the functional form of $f_i(u_i)$, we employ NNETs (universal approximators, provided they have sufficiently many layers and neurons) to approximate the modular functions. If the functional forms are known a priori, then, given the same training data, a traditional nonlinear parametric regression approach can also yield accurate approximations.

Our work, idealizes perfect measurement of global input/output data. In real experimental settings, measurements are often noisy, which may cause the learned functions and parameters to deviate from their true values. Future work will analyze the impact of noise and unmodeled dynamics on the ability to learn the modules' functions.
As a direction for future work, one could explore the possibility of learning only the context-dependent parameters while keeping the previously learned modules fixed. Furthermore, it would be of interest to investigate modular identifiability under more general models that capture additional contextual effects. 

In addition, while this paper focuses on the steady-state input/output mapping, extending the framework to a dynamical systems setting could provide deeper insights into the system’s behavior over time-series input/output data.

\bibliographystyle{IEEEtran}
\bibliography{IEEEabrv, reference}

@book{delvecchio2014biomolecular,
  author    = {Del Vecchio, Domitilla and Murray, Richard M.},
  title     = {Biomolecular Feedback Systems},
  publisher = {Princeton University Press},
  year      = {2014},
  address   = {Princeton, NJ, USA}
}

@article{eslami2022prediction,
  author    = {Eslami, Mohammed and Borujeni, Ali E. and Eramian, Hamid and Weston, Michael and Zheng, Gang and Urrutia, Jennifer and Corbet, Cara and Becker, Daniel and Maschhoff, Kevin and Clowers, Ann and Cristofaro, Angelo and Hosseini, Hamed D. and Gordon, Daniel B. and Dorfan, Yelena and Singer, Julie and Vaughn, Michael and Gaffney, Nicholas and Fonner, Jenna and Stubbs, Christopher A. Voigt and Yeung, Enoch},
  title     = {Prediction of whole‑cell transcriptional response with machine learning},
  journal   = {Bioinformatics},
  volume    = {38},
  number    = {2},
  pages     = {404--409},
  year      = {2022},
  doi       = {10.1093/bioinformatics/btab676}
}

@article{palacios2025machine,
  author    = {Palacios, Sebastian and Collins, James J. and Del Vecchio, Domitilla},
  title     = {Machine learning for synthetic gene circuit engineering},
  journal   = {Current Opinion in Biotechnology},
  volume    = {92},
  pages     = {103263},
  year      = {2025},
  doi       = {10.1016/j.copbio.2025.103263}
}

@article{jayanthi2013retroactivity,
  author    = {S. Jayanthi and K. S. Nilgiriwala and D. Del Vecchio},
  title     = {Retroactivity controls the temporal dynamics of gene transcription},
  journal   = {ACS Synthetic Biology},
  volume    = {2},
  number    = {8},
  pages     = {431--441},
  year      = {2013},
  doi       = {10.1021/sb300098w}
}

@article{grunberg2020modular,
  author    = {T. W. Grunberg and D. Del Vecchio},
  title     = {Modular analysis and design of biological circuits},
  journal   = {Current Opinion in Biotechnology},
  volume    = {63},
  pages     = {41--47},
  year      = {2020},
  doi       = {10.1016/j.copbio.2019.11.015}
}

@article{mishra2014loaddriver,
  author    = {D. Mishra and P. M. Rivera and A. Lin and D. Del Vecchio and R. Weiss},
  title     = {A load driver device for engineering modularity in biological networks},
  journal   = {Nature Biotechnology},
  volume    = {32},
  number    = {12},
  pages     = {1268--1275},
  year      = {2014},
  doi       = {10.1038/nbt.3044}
}

@article{delvecchio2008modular,
  author    = {D. Del Vecchio and A. J. Ninfa and E. D. Sontag},
  title     = {Modular cell biology: retroactivity and insulation},
  journal   = {Molecular Systems Biology},
  volume    = {4},
  number    = {1},
  pages     = {161},
  year      = {2008},
  doi       = {10.1038/msb4100204}
}

@article{qian2017resource,
  author    = {Y. Qian and H.-H. Huang and J. I. Jiménez and D. Del Vecchio},
  title     = {Resource competition shapes the response of genetic circuits},
  journal   = {ACS Synthetic Biology},
  volume    = {6},
  number    = {7},
  pages     = {1263--1272},
  year      = {2017},
  doi       = {10.1021/acssynbio.6b00361}
}

@article{diblasi2024resourcecomp,
  author    = {R. Di Blasi and J. Gabrielli and K. Shabestary and I. Ziarti and others},
  title     = {Understanding resource competition to achieve predictable synthetic gene expression in eukaryotes},
  journal   = {Nature Reviews Bioengineering},
  volume    = {2},
  number    = {9},
  pages     = {721--732},
  year      = {2024},
  doi       = {10.1038/s44222-024-00206-0}
}

@article{diblasi2023resourceaware,
  author    = {R. Di Blasi and M. Pisani and F. Tedeschi and M. M. Marbiah and K. Polizzi and S. Furini and V. Siciliano and F. Ceroni},
  title     = {Resource-aware construct design in mammalian cells},
  journal   = {Nature Communications},
  volume    = {14},
  number    = {1},
  pages     = {3576},
  year      = {2023},
  doi       = {10.1038/s41467-023-39252-4}
}

@article{delvecchio2015modularity,
  author    = {D. Del Vecchio},
  title     = {Modularity, Context-Dependence, and Insulation in Engineered Biological Circuits},
  journal   = {Trends in Biotechnology},
  volume    = {33},
  number    = {2},
  pages     = {111--119},
  month     = {Feb},
  year      = {2015},
  doi       = {10.1016/j.tibtech.2014.11.009}
}

@article{gyorgy2015isocost,
  author    = {A. György and J. I. Jiménez and J. Yazbek and H.-H. Huang and H. Chung and R. Weiss and D. Del Vecchio},
  title     = {Isocost Lines Describe the Cellular Economy of Genetic Circuits},
  journal   = {Biophysical Journal},
  volume    = {109},
  number    = {3},
  pages     = {639--646},
  year      = {2015},
  doi       = {10.1016/j.bpj.2015.06.034}
}

@article{huang2018quasiintegral,
  author    = {H.-H. Huang and Y. Qian and D. Del Vecchio},
  title     = {A quasi-integral controller for adaptation of genetic modules to variable ribosome demand},
  journal   = {Nature Communications},
  volume    = {9},
  pages     = {5415},
  year      = {2018},
  doi       = {10.1038/s41467-018-07899-z}
}

@article{jones2020feedforward,
  author    = {R. D. Jones and Y. Qian and V. Siciliano and B. DiAndreth and J. Huh and R. Weiss and D. Del Vecchio},
  title     = {An endoribonuclease-based feedforward controller for decoupling resource-limited genetic modules in mammalian cells},
  journal   = {Nature Communications},
  volume    = {11},
  pages     = {5690},
  year      = {2020},
  doi       = {10.1038/s41467-020-19126-9}
}

@article{frei2020mitigation,
  author    = {T. Frei and F. Cella and F. Tedeschi and J. Gutiérrez and G.-B. Stan and M. Khammash and V. Siciliano},
  title     = {Characterization and mitigation of gene expression burden in mammalian cells},
  journal   = {Nature Communications},
  volume    = {11},
  pages     = {4641},
  year      = {2020},
  doi       = {10.1038/s41467-020-18392-x}
}

@article{pandey2023bioscrape,
  author    = {A. Pandey and W. Poole and A. Swaminathan and V. Hsiao and R. M. Murray},
  title     = {Fast and flexible simulation and parameter estimation for synthetic biology using bioscrape},
  journal   = {Journal of Open Source Software},
  volume    = {8},
  number    = {83},
  pages     = {5057},
  year      = {2023},
  doi       = {10.21105/joss.05057}
}

@article{poole2022biocrnpyler,
  author    = {W. Poole and A. Pandey and A. Shur and Z. A. Tuza and R. M. Murray},
  title     = {BioCRNpyler: Compiling chemical reaction networks from biomolecular parts in diverse contexts},
  journal   = {PLoS Computational Biology},
  volume    = {18},
  number    = {4},
  pages     = {e1009987},
  year      = {2022},
  doi       = {10.1371/journal.pcbi.1009987}
}

@article{karniadakis2021physicsinformed,
  author    = {G. E. Karniadakis and I. G. Kevrekidis and L. Lu and P. Perdikaris and S. Wang and L. Yang},
  title     = {Physics-informed machine learning},
  journal   = {Nature Reviews Physics},
  volume    = {3},
  number    = {6},
  pages     = {422--440},
  year      = {2021},
  doi       = {10.1038/s42254-021-00314-5}
}

@article{darabi2025combining,
  author    = {A. Darabi and Z. An and M. A. Al-Radhawi and W. Cho and M. Siami and E. D. Sontag},
  title     = {Combining model-based and data-driven models: an application to synthetic biology resource competition},
  journal   = {Mathematical Biosciences},
  year      = {2026}
}

@book{hungerford1974algebra,
  author       = {Hungerford, Thomas W.},
  title        = {{Algebra}},
  series       = {Graduate Texts in Mathematics},
  volume       = {73},
  publisher    = {Springer-Verlag},
  address      = {New York},
  year         = {1974},
  isbn         = {0-387-90518-9},
  note         = {Volume~73 in GTM series},
}

@article{alcantar2024highthroughput,
  author    = {M. A. Alcantar and M. A. English and J. A. Valeri and J. J. Collins},
  title     = {A High-Throughput Synthetic Biology Approach for Studying Combinatorial Chromatin-Based Transcriptional Regulation},
  journal   = {Molecular Cell},
  volume    = {84},
  number    = {12},
  pages     = {2382--2396.e9},
  year      = {2024},
  doi       = {10.1016/j.molcel.2024.05.025}
}

@article{dankers2016predictorinput,
  author    = {A. Dankers and P. M. J. Van den Hof and X. Bombois and P. S. C. Heuberger},
  title     = {Identification of dynamic models in complex networks with Prediction Error Methods: Predictor Input Selection},
  journal   = {IEEE Transactions on Automatic Control},
  volume    = {61},
  number    = {4},
  pages     = {937--952},
  year      = {2016},
  doi       = {10.1109/TAC.2015.2450895}
}

@article{vandenhof2013consistentmodule,
  author    = {P. M. J. van den Hof and A. G. Dankers and P. S. C. Heuberger and X. J. A. Bombois},
  title     = {Identification of Dynamic Models in Complex Networks with Prediction Error Methods: Basic Methods for Consistent Module Estimates},
  journal   = {Automatica},
  volume    = {49},
  number    = {10},
  pages     = {2994--3006},
  year      = {2013},
  doi       = {10.1016/j.automatica.2013.07.011}
}

@book{munkres2000topology,
  author    = {James R. Munkres},
  title     = {Topology},
  edition   = {2nd},
  publisher = {Prentice Hall},
  year      = {2000},
  address   = {Upper Saddle River, NJ, USA},
  pages     = {537},
  isbn      = {978-0131816299}
}

@inproceedings{vizuete2023nonlinear,
  title={Nonlinear network identifiability: The static case},
  author={Vizuete, Renato and Hendrickx, Julien M},
  booktitle={2023 62nd IEEE Conference on Decision and Control (CDC)},
  pages={443--448},
  year={2023},
  organization={IEEE}
}

@article{vizuete2024nonlinear,
  title={Nonlinear network identifiability with full excitations},
  author={Vizuete, Renato and Hendrickx, Julien M},
  journal={arXiv preprint arXiv:2405.07636},
  year={2024}
}

@inproceedings{he2015delving,
  title={Delving deep into rectifiers: Surpassing human-level performance on imagenet classification},
  author={He, Kaiming and Zhang, Xiangyu and Ren, Shaoqing and Sun, Jian},
  booktitle={Proceedings of the IEEE international conference on computer vision},
  pages={1026--1034},
  year={2015}
}

@article{sontag1979finitary,
  title={On Finitary Linear Systems},
  author={Sontag, Eduardo D.},
  journal={Kybernetika},
  volume={15},
  number={5},
  pages={350--358},
  year={1979}
}

\end{document}